\documentclass[runningheads]{llncs}
\usepackage[T1]{fontenc}
\usepackage{graphicx}

\usepackage{booktabs}
\usepackage{multirow}
\usepackage{array}

\newcommand{\ms}[2]{$#1_{(#2)}$}
\newcommand{\bms}[2]{$\mathbf{#1}_{(#2)}$}
\newcommand{\sbms}[2]{$\underline{#1}_{(#2)}$}

\usepackage{newtxmath}
\usepackage{color}
\usepackage{algorithmic,algorithm}
\usepackage{subcaption}
\begin{document}
\def\method{TPET}
\title{Evolutionary Discovery of Heuristic Policies for Traffic Signal Control}
%
%
\author{
Ruibing Wang\inst{1} \and  
Shuhan Guo\inst{2} \and   
Zeen Li\inst{2} \and        
Zhen Wang\inst{1} \and    
Quanming Yao\inst{2}      
}

\authorrunning{R. Wang et al.} 


\institute{
Northwestern Polytechnical University, Xi'an, China\\
\email{wrb5261@mail.nwpu.edu.cn, w-zhen@nwpu.edu.cn} 
\and
Tsinghua University, Beijing, China\\
\email{guoshuhan@tsinghua.edu.cn, lze25@mails.tsinghua.edu.cn, qyaoaa@tsinghua.edu.cn}
}


%
\maketitle              
\begin{abstract}
Traffic Signal Control (TSC) involves a challenging trade-off: classic heuristics are efficient but oversimplified, while Deep Reinforcement Learning (DRL) achieves high performance yet suffers from poor generalization and opaque policies. Online Large Language Models (LLMs) provide general reasoning but incur high latency and lack environment-specific optimization. To address these issues, we propose Temporal Policy Evolution for Traffic (\textbf{\method{}}), which uses LLMs as an evolution engine to derive specialized heuristic policies. The framework introduces two key modules: (1) Structured State Abstraction (SSA), converting high-dimensional traffic data into temporal-logical facts for reasoning; and (2) Credit Assignment Feedback (CAF), tracing flawed micro-decisions to poor macro-outcomes for targeted critique. Operating entirely at the prompt level without training, \method{} yields lightweight, robust policies optimized for specific traffic environments, outperforming both heuristics and online LLM actors.

\end{abstract}
\section{Introduction}
Traffic Signal Control (TSC)~\cite{shelby2004single} is a canonical sequential decision-making problem at the heart of urban mobility. The efficient dispatching of signal phases directly impacts commuter travel time, fuel consumption, and public safety. This high-stakes domain demands policies that are not only efficient but also robust to the non-stationary, dynamic nature of traffic flows.

In striving for effective traffic signal control, researchers have encountered a fundamental trade-off between generality and specialization. Classic transportation models, such as Fixed-Time~\cite{fixedtime} and MaxPressure~\cite{maxpressure}, represent the general-purpose solution. These methods function as traditional heuristic policies: they rely on manually designed, explicit rules to map observed traffic states directly to signal actions, without the need for parameter training. While computationally trivial and fast, their reliance on simple, myopic logic makes them fundamentally sub-optimal for any specific intersection's unique topology and complex flow patterns. At the other end of the spectrum, Deep Reinforcement Learning (DRL) methods offer a path to deep specialization and can achieve high performance. This performance, however, comes at a significant cost: DRL agents are notorious "black boxes" with unverifiable policies, and their intensive training on specific data distributions often leads to brittle policies with poor generalization. Recently, Large Language Models (LLMs) have been proposed as a new paradigm, using a general-purpose LLM as an online actor. This approach, while intellectually appealing, suffers from critical, real-world flaws, including prohibitive inference latency incompatible with high-frequency decisions and a fundamental lack of environment-specific specialization.

To address these limitations, We argue that the optimal role for an LLM is not as a slow, generic online actor, but as a powerful discoverer. We introduce the Temporal Policy Evolution for Traffic (\textbf{\method{}}) framework, which leverages an LLM-driven evolutionary engine to evolve a specialized, structured, and lightweight policy from the ground up. The final output is a simple heuristic function that solves both problems of the online LLM: it runs with millisecond latency, and it is deeply specialized, having been optimized over thousands of simulated generations specifically for the target environment.

Our scientific hypothesis is that this specialization will lead to a superior performance compared to the generic reasoning of an online LLM and the simplistic logic of classic heuristics. To achieve this, we introduce two key modules to empower the LLM in this discovery process: the Structured State Abstraction (SSA) module to bridge the numerical-to-Semantic gap, and the Credit Assignment Feedback (CAF) module to provide rich, explanatory critiques, allowing the LLM to perform targeted, intelligent mutations that adapt the policy to the environment's unique challenges. 

Our contributions can be summarized as follows:
\vspace{-0.2cm}
\begin{itemize}
	\item We introduce the Evolutionary Discovery paradigm for TSC, which leverages an LLM to discover specialized structured policies, offering a novel, practical alternative to fixed strategies and online LLM actors.
	
	\item We design and implement the \method{} framework with two novel modules: SSA, a state abstraction method for high-frequency time-series data, and CAF, a credit assignment mechanism that back-traces simulation logs for critique-driven evolution.

	\item We demonstrate through experiments in CityFlow that our discovered policy achieves state-of-the-art performance advantage over both classic heuristics of transportation models and the general-purpose online LLM actor.
\end{itemize}

\section{Related Work}

\subsection{Traffic Signal Control (TSC) Problem}
The pursuit of effective TSC strategies has broadly followed three paradigms. The most traditional class, classic transportation strategies, provides general-purpose, computationally trivial solutions such as Fixed-Time~\cite{fixedtime} and MaxPressure series~\cite{maxpressure,efficient_pressure,summarymaxpressure}. While fast, their simple, myopic logic is sub-optimal for complex flows. To achieve deep specialization, Reinforcement Learning (RL) based methods emerged, with a large body of work including MPLight~\cite{mplight}, AttendLight~\cite{attendlight}, PressLight~\cite{presslight}, and advanced methods~\cite{advancedxlight,gplight+}. These methods achieve state-of-the-art performance in simulations but are notorious "black boxes" with unverifiable policies and poor generalization. Most recently, Online LLM-based Approaches have been proposed~\cite{dualllm,llmlight,collm,crossroadllm,trafficai}, which use a general LLM as a decision-maker. This paradigm, however, suffers from prohibitive latency and a lack of specialization.


\subsection{LLM for Heuristic Evolution}
Recent advancements have established Large Language Models (LLMs) as engines for automated heuristic design, marking a shift towards "Language Hyper-Heuristics" where LLMs act as evolutionary operators within function space. These processes typically build on existing heuristics~\cite{tabu,ACO,pso} and aim to refine them automatically. FunSearch~\cite{funsearch} couples a pre-trained LLM with a systematic evaluator to discover new mathematical knowledge and algorithms. EoH~\cite{eoh} extends this idea by co-evolving natural language thoughts and executable codes, while ReEvo~\cite{reevo} introduces reflection to generate verbal gradients from past performance, improving efficiency and interpretability. NeRM~\cite{nerm} further addresses misalignment between task descriptions and solutions through nested refinement of prompts and algorithms, assisted by predictor-based evaluation. Collectively, these works illustrate a progression from stochastic code search to reflective and structured evolutionary frameworks.

\section{Methodology}

\begin{figure}[htb]
	\centering
	\includegraphics[width=\textwidth]{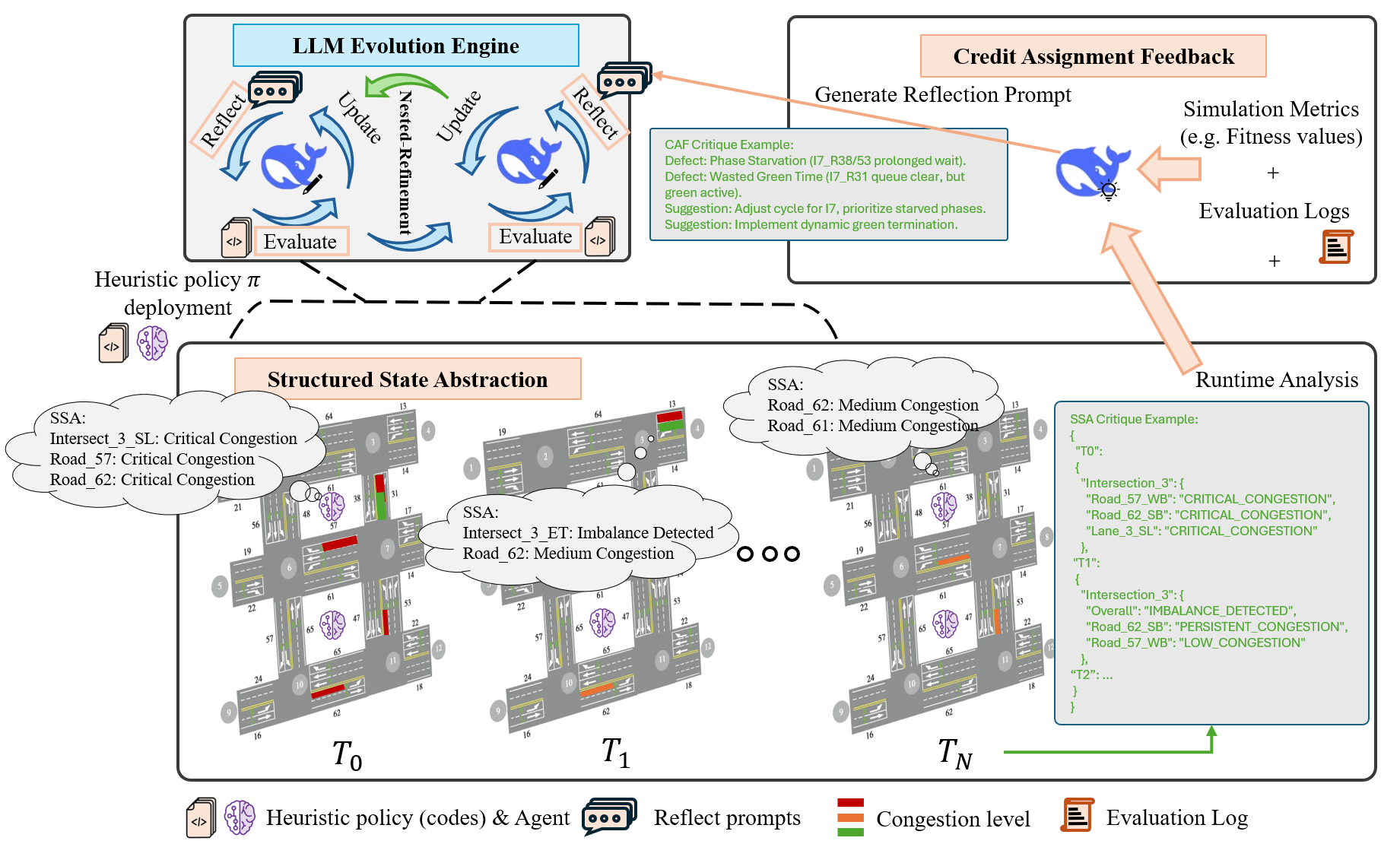}
	\caption{Framework of \method{}, depicting an LLM iteratively evolving heuristic traffic control policies by receiving actionable defect feedback from the Credit Assignment Feedback (CAF) module, which analyzes real-time traffic conditions abstracted by the Structured State Abstraction (SSA) module.} \label{fig::framework}
    \vspace{-0.4cm}
\end{figure}
\vspace{-0.2cm}

\subsection{Problem Formulation}
Traffic Signal Control is a canonical sequential decision-making problem. Following LLMLight~\cite{llmlight}, we model the urban traffic network as a graph $G = (V, E)$, where each intersection $v \in V$ is formulated as a discrete-time Markov Decision Process (MDP). The objective is to minimize total waiting time, queue length, and travel time of vehicles. At each step $t$, the agent observes a state $s_t \in S$, represented by a high-dimensional vector that captures the traffic situation, including queue length (the number of waiting vehicles for each movement) and waiting time (the accumulated delays reflecting temporal pressure and fairness).
The action $a_t \in A$ is the selection of the next signal phase from a predefined set, where each phase corresponds to a group of non-conflicting traffic movements given green simultaneously. The agent’s decision must respect safety constraints such as minimum green times and clearance intervals.

\subsection{Foundational Framework}

Our framework builds upon the biologically inspired co-evolutionary paradigm introduced in NeRM~\cite{nerm}. 
The core idea is to jointly evolve task specifications and algorithmic solutions through a nested loop of two interleaved modules: 
Metamorphosis on Prompts (MoP), which iteratively refines natural language task descriptions, and Metamorphosis on Algorithms (MoA), which evolves code solutions conditioned on these refined prompts. 
While NeRM originally utilizes LLMs to evolve heuristic code for combinatorial optimization, we adapt this methodology to the domain of sequential decision-making. 
Specifically, the evolutionary engine in \method{} targets the generation of heuristic policies for traffic control, optimizing the core decision-making logic (e.g., pressure calculation rules) encapsulated within the control loop.
 

\vspace{-0.4cm}
\subsubsection{Challenges in Adaptation} 
However, directly applying this general framework to Traffic Signal Control faces two distinct challenges. First, the Semantic Gap hinders reasoning, as traffic simulators output high-dimensional numerical data while LLMs rely on semantic logic. Second, since rewards in this sequential domain are sparse and delayed, the standard scalar fitness score fails to identify the specific micro-decisions causing macro-level failures.
\vspace{-0.3cm}
\subsection{Temporal Policy Evolution for Traffic (\method{})}

To overcome the challenges identified above, we propose the TPET framework, which equips the foundational co-evolutionary backbone with two specific mechanisms. We introduce Structured State Abstraction (SSA) to bridge the semantic gap by translating numerical states into interpretable logical facts , and Credit Assignment Feedback (CAF) to resolve the credit assignment problem by replacing sparse rewards with dense, actionable critiques derived from simulation logs.
\vspace{-0.4cm}
\subsubsection{Structured State Abstraction (SSA) for Context Translation}
This module is developed to overcome the semantic gap. The raw state vector $s_t$ is numerical, while the heuristic $\pi$ we aim to discover should operate on robust, logical concepts. Furthermore, $s_t$ is an instantaneous snapshot, lacking the crucial temporal history $H_{t-1}$ required for sequential decisions. The SSA module is designed as a formal, deterministic function $f_{SSA}: (S_t, H_{t-1}) \rightarrow \Sigma^*_t$. Its purpose is to act as the input interface for the policy $\pi$. During online validation, $\pi$ calls SSA at each step $t$ to translate the raw numerical state $s_t$ and its history $H_{t-1}$ into a discrete, structured alphabet $\Sigma^*_t$.

This transformation is a knowledge-driven synthesis pipeline designed to extract salient temporal-logical facts. This process involves two stages:
\vspace{-0.2cm}
\begin{itemize}
	\item Metric Aggregation and Persistence: The module computes instantaneous aggregate metrics (such as maximum pressure). Crucially, it also maintains and updates persistent temporal metrics (such as a starvation timer for each phase), which explicitly encode the system's history.
	\item Temporal Predicate Generation: The module applies a set of logical rules to map these quantitative metrics (both instantaneous and persistent) onto a stable, qualitative vocabulary, generating structured facts (predicates).
\end{itemize}
\vspace{-0.2cm}

The resulting alphabet $\Sigma^*_t$ is a structured set of facts describing the complete state. 
This set includes the following predicates:
\vspace{-0.2cm}
\begin{itemize}
    \item Congestion Predicates: These describe instantaneous pressure. 
    \begin{itemize}
        \item Examples: ``Congestion: Critical'', ``Congestion: High'', ``Congestion: Moderate'', ``Congestion: Low'', or ``Dominant Flow: Phase 2''.
    \end{itemize}

    \item Temporal-Fairness Predicates: These capture historical context. 
    \begin{itemize}
        \item Examples include ``Starvation Risk: High'' (triggered by $\text{pressure} > \theta_{demand}$ and $\text{starvation\_timer} > \tau_{critical}$) or ``Queue Urgency: Critical'' (derived from \texttt{max\_wait\_time}).
    \end{itemize}

    \item Balance Predicates: These describe relationships between flows.
    \begin{itemize}
        \item Examples: ``Imbalance: EW Dominant'' or ``Imbalance: None''.
    \end{itemize}
\end{itemize}
\vspace{-0.2cm}







\begin{figure}[h!]
        \vspace{-0.5cm}
	\centering
	
	\begin{subfigure}{0.8\textwidth}
		\centering
		\includegraphics[width=\linewidth]{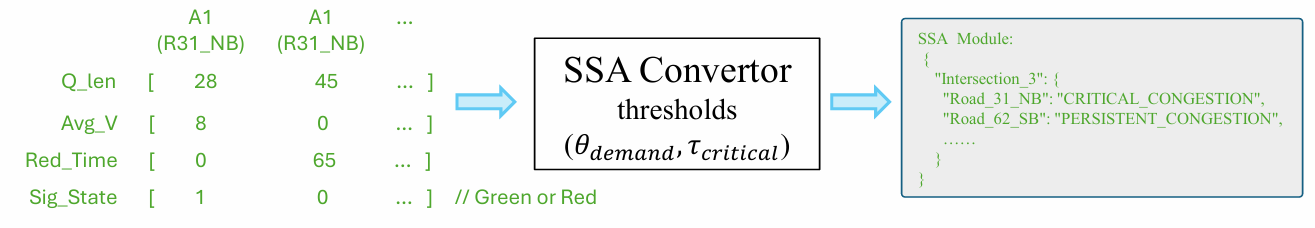} 
		\caption{Process of SSA module.}
		\label{fig:ssa}
	\end{subfigure}
	
	\vspace{1em}
	
	\begin{subfigure}{0.8\textwidth}
		\centering
		\includegraphics[width=\linewidth]{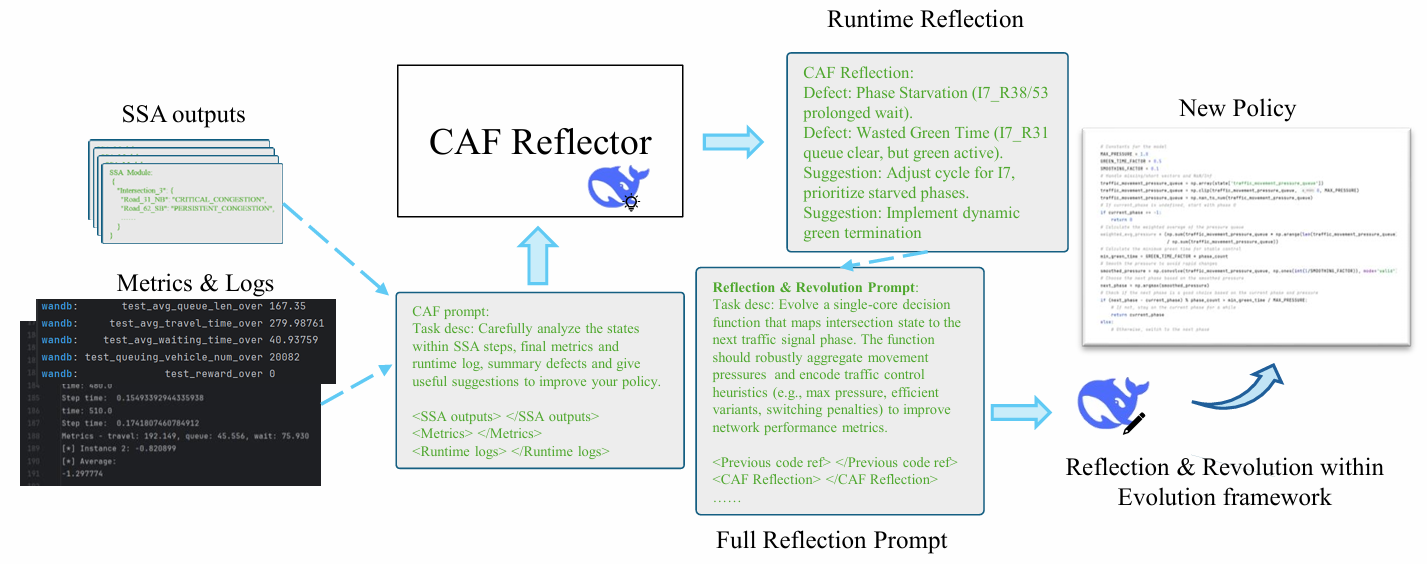}
		\caption{Process of CAF module.}
		\label{fig:caf}
	\end{subfigure}

	\caption{Details design of key modules.} 
	\label{fig:module_details}
        \vspace{-0.8cm}
\end{figure}

It is critical to understand how this module functions within the \method{} loop. During online validation, the evolved policy $\pi_i$ (the Python function) calls the $f_{SSA}$ function at every step $t$ to acquire its numerical input $\Sigma^*_t$. The logic within $\pi_i$ then operates on these inputs. During the evolution stage, the LLM is provided with the description of the structured vocabulary that $f_{SSA}$ produces. This defined interface is the crucial information support that allows the LLM to write a new policy $\pi_{child}$ that intelligently utilizes these pre-defined structured concepts to address the flaws of the previous policy. An example has been shown in Fig.~\ref{fig:ssa}.

\subsubsection{Credit Assignment Feedback (CAF) for Post-Hoc Defect Analysis} \label{sec::caf}
This module aims to address the temporal credit assignment problem. When a policy $\pi_i$ completes the evaluation, its final fitness score is an intractably sparse reward. 

The CAF module is a post-hoc analysis engine designed to solve this. After the simulation is complete, it performs heuristic-driven back-tracing on the entire simulation log $L = \{(\Sigma^*_0, a_0), (\Sigma^*_1, a_1), ..., (\Sigma^*_T, a_T)\}$. The core of the CAF is a Defect Pattern Library containing formal definitions of common failure patterns. These patterns are expressed as temporal-logical rules that connect flawed micro-decisions to negative macro-outcomes:
\vspace{-0.3cm}
\begin{itemize}
	\item Wasted Green Time: A spatial-instantaneous failure, defined as the agent selecting a phase $a_t = i$ when its corresponding structured state $\Sigma^*_t$ indicated near-zero demand.
	\item Phase Starvation: A severe temporal failure. This pattern is matched if the log $L$ shows a phase $i$ that has not been activated for a critical duration $\tau_{crit}$ (i.e., $a_k \neq i$ for $k \in [t-\tau_{crit}, t]$) and its structured state $\Sigma^*_t$ indicated high demand.
	\item Premature Phase Switch: A temporal inefficiency. This pattern is matched if the agent was serving a high-demand phase $i$ (indicated by $\Sigma^*_{t-1}$) but switched to phase $j$ ($a_t = j$), incurring the time cost of a yellow-light cycle without fully serving the demand.
\end{itemize}
\vspace{-0.3cm}

The CAF module scans the log, aggregates these defects, and generates a structured critique $C_i$ for the entire policy $\pi_i$. This critique is the primary feedback signal for the LLM. This process replaces the simple fitness score of a standard evolutionary algorithm with a rich, actionable debug report. An example has been shown in Fig.~\ref{fig:caf}. Utilizing these two modules, the process is shown in Algorithm~\ref{alg_tpet}.


\begin{algorithm}[ht]
	\caption{Learning Process of \method{}}
	\label{alg_tpet}
    
	\noindent \textbf{Input:} Problem description; initial algorithms; LLM; \\
	\textbf{Output}: Generated optimal algorithm.
    
	\begin{algorithmic}[1]
		\STATE $h^\star \leftarrow$ initial algorithm
		\WHILE{true}
		\STATE // Prompt \& Algorithm Evolution with SSA \& CAF
		\FOR{$j=1,2,\cdots,T$}
		\FOR{$k=1,2,\cdots,T$}
		\STATE ...
		\STATE Reflection \& Revolution (with CAF prompt)
		\STATE ...
		\STATE Evaluation new policy with Real-time Analysis (with SSA module)
		\ENDFOR
		\ENDFOR
		\STATE $h^\star \leftarrow$ best algorithm from current evolution
		\STATE Deploy $h^\star$ in intersections
		\ENDWHILE
	\end{algorithmic}
\end{algorithm}
\vspace{-0.4cm}
\subsection{Comparison with Previous Works}

While our evolutionary loop is methodologically based on NeRM's MoP and MoA components , \method{} successfully adapts this paradigm to a fully dynamic, high-frequency, sequential decision-making domain. This is a non-trivial leap, as NeRM was designed for static problems using a single fitness score, and NeRM-Net addressed static batch allocation. Our adaptation is enabled exclusively by two modules: SSA serves as the structured interface for the policy, and CAF replaces the simple feedback of prior work with a sequential critique. This reframes the LLM's task from finding a static heuristic to debugging the temporal-logical flaws in a dynamic policy. The final discovered policy is an interpretable, robust algorithm, evolved to explicitly handle the temporal defects identified by CAF.

\section{Experiments}
\subsection{Experimental Settings}
\subsubsection{Datasets}
Our experiments use three real-world traffic flow datasets. Specifically, we select two datasets from Jinan and one dataset from Hangzhou. 

\begin{itemize}
    \item \textbf{Jinan-1 \& Jinan-2:} These two datasets is collected from Dongfeng sub-district, Jinan, China, featuring 12 intersections. The covered area is approximately 400 meters east-west and 800 meters north-south.
    
    \item \textbf{Hangzhou:} This dataset is collected from Gudang sub-district, Hangzhou, China, featuring 16 intersections. The covered area spans 800 meters east-west and 600 meters north-south.
\end{itemize}

\subsubsection{Evaluation Metrics}
Following previous studies, we evaluate performance using three standard metrics. These are: Average Travel Time (ATT), the average time for a vehicle to travel from entry to exit; Average Queue Length (AQL), the average number of stationary vehicles across all lanes; and Average Wait Time (AWT), the average accumulated time vehicles spend in a stationary state.

\vspace{-0.4cm}
\subsubsection{Implementation Details}
All experiments are conducted within the CityFlow~\cite{zhang2019cityflow} simulation environment. The LLM Evolution Engine  is implemented using the GLM-4-Flash model as its core. Our computational environment is equipped with an NVIDIA RTX 3070Ti GPU, an Intel i7-11700K CPU, and 32GB of RAM. During each iteration of the evolutionary loop, 20 policy candidates are generated. Following the critique from the CAF module and the overall fitness evaluation, the top 3 candidates are retained for the next generation. To assess the performance and stability of the discovered policies, each experiment is run three times and has 20 iterations, with the mean and standard deviation reported in our results.

\begin{table*}[t]
\vspace{-10pt}
\centering
\caption{Overall performance comparison of \method{} against traditional transportation, reinforcement learning, and LLM-enhanced methods on the Jinan and Hangzhou datasets.
The best results are bolded, the second-best results are underlined. }
\label{tab:mainresults}
\resizebox{\textwidth}{!}{
\begin{tabular}{l|ccc|ccc|ccc}
\hline
\multirow{2}{*}{Models} & \multicolumn{3}{c|}{Jinan-1} & \multicolumn{3}{c|}{Jinan-2} & \multicolumn{3}{c}{Hangzhou} \\
 & ATT & AQL & AWT & ATT & AQL & AWT & ATT & AQL & AWT \\
\hline
\multicolumn{10}{c}{\textbf{Transportation Methods}} \\
\hline
Random & \ms{552.74}{12.7} & \ms{529.63}{20.05} & \ms{99.33}{9.18} & \ms{555.23}{13.72} & \ms{428.38}{19.77} & \ms{100.40}{5.64} & \ms{621.14}{20.37} & \ms{295.81}{17.08} & \ms{96.06}{6.46} \\
FixedTime & \ms{450.11}{0.00} & \ms{394.34}{0.00} & \ms{69.19}{0.00} & \ms{441.19}{0.00} & \ms{294.14}{0.00} & \ms{66.72}{0.00} & \ms{616.02}{0.00} & \ms{301.33}{0.00} & \ms{73.99}{0.00} \\
Maxpressure & \ms{265.75}{0.00} & \ms{133.90}{0.00} & \sbms{40.20}{0.00} & \ms{273.20}{0.00} & \ms{106.58}{0.00} & \sbms{38.25}{0.00} & \ms{325.33}{0.00} & \ms{68.99}{0.00} & \sbms{49.60}{0.00} \\
\hline
\multicolumn{10}{c}{\textbf{RL Methods}} \\
\hline
MPLight & \ms{291.79}{1.25} & \ms{171.70}{0.92} & \ms{89.93}{2.40} & \ms{304.51}{0.88} & \ms{142.25}{0.74} & \ms{90.91}{1.93} & \ms{345.60}{0.96} & \ms{84.70}{0.83} & \ms{81.97}{2.79} \\
AttendLight & \ms{273.02}{0.87} & \ms{144.05}{0.74} & \ms{55.93}{2.12} & \ms{280.94}{0.66} & \ms{115.52}{0.93} & \ms{52.46}{1.67} & \ms{322.94}{0.78} & \ms{66.96}{0.55} & \ms{55.19}{1.41} \\
PressLight & \ms{275.85}{0.74} & \ms{148.18}{0.81} & \ms{54.81}{1.39} & \ms{281.46}{0.75} & \ms{115.99}{0.62} & \ms{47.27}{2.38} & \ms{364.13}{0.67} & \ms{98.67}{0.73} & \ms{90.33}{1.40} \\
CoLight & \ms{266.39}{0.60} & \ms{135.08}{0.82} & \ms{53.33}{2.02} & \ms{274.77}{0.54} & \ms{108.28}{0.72} & \ms{54.14}{1.39} & \ms{322.85}{0.65} & \ms{66.94}{0.61} & \ms{61.82}{2.38} \\
\hline
\multicolumn{10}{c}{\textbf{LLMLight (with Generalist LLMs)}} \\
\hline
Qwen2-72B & \ms{274.33}{12.72} & \ms{144.95}{13.70} & \ms{54.94}{3.68} & \ms{277.78}{13.74} & \ms{112.13}{11.71} & \ms{53.12}{4.69} & \ms{321.91}{14.73} & \ms{65.75}{12.70} & \ms{66.52}{5.68} \\
Llama2-70B & \ms{320.41}{12.80} & \ms{210.13}{11.78} & \ms{100.90}{2.75} & \ms{324.52}{13.82} & \ms{162.34}{12.79} & \ms{99.87}{4.76} & \ms{357.95}{10.81} & \ms{93.21}{11.77} & \ms{106.63}{2.74} \\
Llama3-70B & \ms{271.60}{7.76} & \ms{142.95}{11.73} & \ms{54.55}{2.70} & \ms{277.49}{10.78} & \ms{112.86}{7.74} & \ms{51.76}{2.71} & \ms{325.85}{8.77} & \ms{69.42}{9.73} & \ms{71.51}{3.70} \\
ChatGPT-3.5 & \ms{501.36}{11.90} & \ms{479.50}{13.88} & \ms{154.19}{4.85} & \ms{524.81}{9.91} & \ms{393.21}{12.87} & \ms{179.34}{2.84} & \ms{463.04}{10.89} & \ms{181.95}{11.86} & \ms{191.87}{3.83} \\
GPT-4 & \bms{264.70}{7.74} & \bms{132.53}{9.71} & \ms{46.16}{1.69} & \bms{271.34}{8.75} & \bms{105.22}{7.72} & \ms{47.55}{2.70} & \sbms{318.71}{9.20} & \sbms{62.84}{7.62} & \ms{58.09}{1.37} \\
\hline
\multicolumn{10}{c}{\textbf{Our Method}} \\
\hline
NeRM & \ms{278.45}{0.54} & \ms{149.20}{0.42} & \ms{55.10}{0.43} & \ms{288.70}{0.49} & \ms{123.60}{0.35} & \ms{54.80}{0.39} & \ms{330.25}{0.59} & \ms{72.10}{0.56} & \ms{62.35}{0.29} \\
\method{} & \sbms{265.58}{0.61} & \sbms{133.43}{0.47} & \bms{36.80}{0.49} & \sbms{271.67}{0.52} & \sbms{105.30}{0.38} & \bms{35.08}{0.42} & \bms{313.64}{0.63} & \bms{59.13}{0.59} & \bms{31.41}{0.28} \\
\hline
\end{tabular}}
\vspace{-10pt}
\end{table*}
\vspace{-0.4cm}
\subsubsection{Compared Models}
To provide a comprehensive evaluation, we consider three categories of comparison methods. First, traditional transportation strategies such as Random, FixedTime~\cite{fixedtime}, and Maxpressure~\cite{maxpressure,efficient_pressure,summarymaxpressure} are included as representative baselines. Second, we benchmark against several reinforcement learning algorithms, including MPLight~\cite{mplight}, AttendLight~\cite{attendlight}, PressLight~\cite{presslight}, and CoLight~\cite{wei2019colight}. Finally, we instantiate LLMLight~\cite{llmlight} with different large language model backbones, specifically GPT-4, ChatGPT-3.5, Qwen, Llama2, and Llama3.

\subsection{Performance Comparison}
\label{sec:performance_comparison}

The experimental results in Table~\ref{tab:mainresults} confirm that our TPET framework delivers strong performance, validating the hypothesis that an LLM-driven evolutionary process guided by SSA and CAF can discover specialized policies that outperform heuristics and rival complex models. \method{} consistently surpasses traditional strategies such as FixedTime and Maxpressure, and establishes itself as highly competitive with state-of-the-art reinforcement learning methods and online LLM-Light actors. On key metrics like Average Travel Time, its performance is superior to or on par with the strongest baselines, while its exceptional results on Average Wait Time highlight the effectiveness of Attribution Feedback in addressing temporal blind spots often overlooked by other paradigms.

A particularly important comparison is with NeRM, which evolves policies using only sparse fitness scores and lacks our feedback modules. \method{} achieves substantial gains over NeRM, demonstrating that the contributions of SSA and CAF are critical for effective policy discovery. 

\begin{figure*}[t!] 
    \centering
    \includegraphics[width=1.0\textwidth]{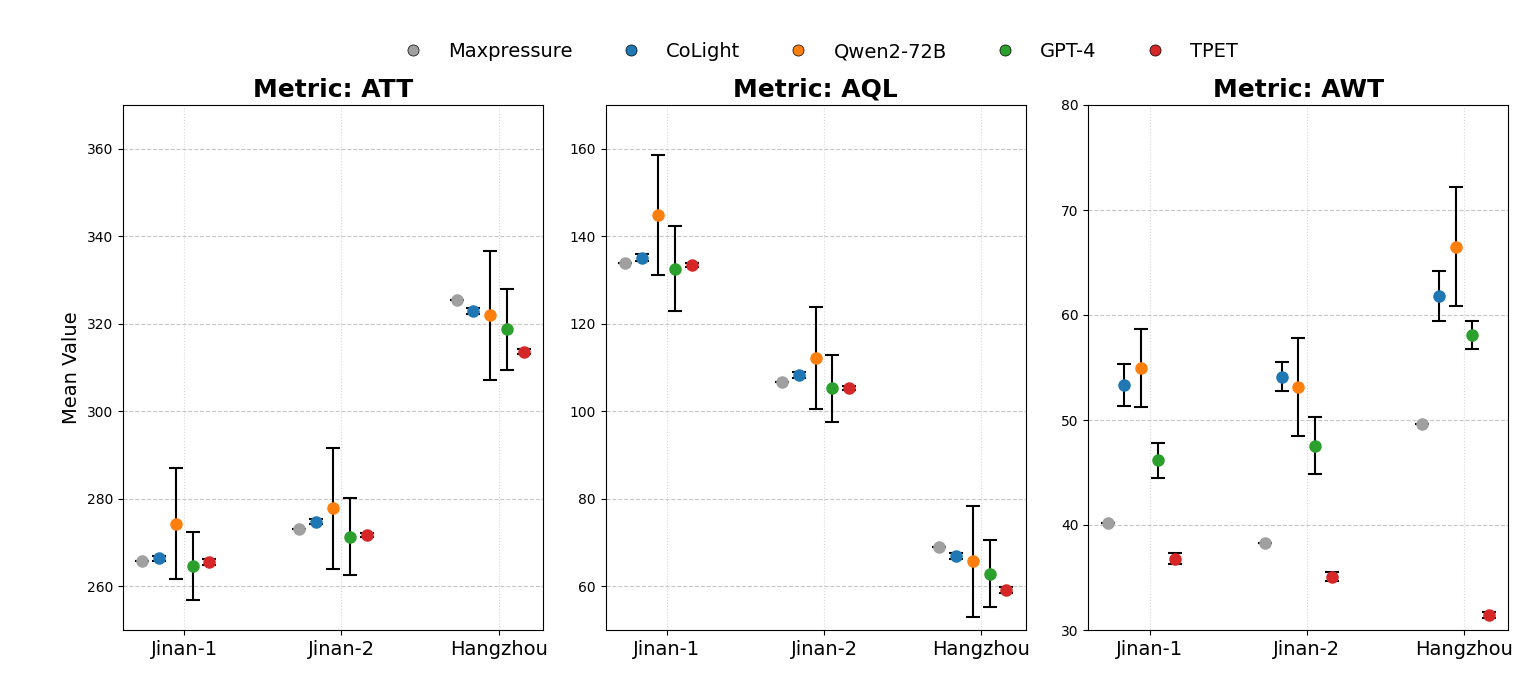} 
    \caption{
    Performance and robustness comparison on key metrics (ATT, AQL, AWT). 
    Lower values are better for all metrics.
    Each point represents the mean value, and the error bar represents the standard deviation (±SD) across multiple runs. 
    Note the Y-axis is zoomed in for each metric to highlight variance. 
    \method{} achieves top-tier performance with minimal variance,
    in stark contrast to the high instability (large error bars) of LLM-based actors.
}
    \label{fig:robustness_plot} 
    \vspace{-10pt} 
\end{figure*}

Beyond performance, \method{} demonstrates exceptional stability and robustness. As shown in Figure~\ref{fig:robustness_plot}, which reports mean performance and standard deviation, generalist LLM actors such as Qwen2-72B and GPT-4 suffer from large error bars, indicating instability. In contrast, \method{} maintains consistently tight variance, often comparable to deterministic methods like Maxpressure. This stability confirms that TPET not only achieves strong performance but also delivers optimized and reliable policies, making it a practical solution for real-world traffic signal control.

\vspace{-2pt}
\subsection{Ablation Study}

To validate the effectiveness of our framework's core components, we conduct an ablation study, with results shown in Table~\ref{tab:ablation_reformatted}. We individually analyze the contribution of our two proposed modules: (1) w/o SSA, which removes the Structured State Abstraction, forcing the policy to operate on less processed information; and (2) w/o CAF, which removes the Credit Assignment Feedback, reverting the evolution to a simpler search.

\begin{table}[t]
    \vspace{-20pt}
    \centering
    \footnotesize
    \caption{Ablation study.}
    \label{tab:ablation_reformatted}
    \resizebox{\textwidth}{!}{
    \begin{tabular}{lccc ccc ccc}
        \toprule
        \multirow{2}{*}{\textbf{Models}} 
        & \multicolumn{3}{c}{\textbf{Jinan-1}} 
        & \multicolumn{3}{c}{\textbf{Jinan-2}} 
        & \multicolumn{3}{c}{\textbf{Hangzhou}} \\
        \cmidrule(lr){2-4} \cmidrule(lr){5-7} \cmidrule(lr){8-10}
        & ATT & AQL & AWT & ATT & AQL & AWT & ATT & AQL & AWT \\
        \midrule
        \method{} & \ms{265.58}{0.61} & \ms{133.93}{0.47} & \ms{36.80}{0.49} & \ms{271.67}{0.52} & \ms{105.30}{0.38} & \ms{35.08}{0.42} & \ms{313.64}{0.63} & \ms{59.13}{0.59} & \ms{31.41}{0.28} \\
        w/o SSA       & \ms{266.53}{0.55} & \ms{135.04}{0.43} & \ms{43.80}{0.48} & \ms{273.08}{0.50} & \ms{107.50}{0.36} & \ms{40.43}{0.41} & \ms{322.67}{0.60} & \ms{66.70}{0.57} & \ms{53.01}{0.32} \\
        w/o CAF       & \ms{275.70}{0.58} & \ms{146.94}{0.46} & \ms{53.26}{0.44} & \ms{285.32}{0.53} & \ms{120.18}{0.39} & \ms{52.55}{0.40} & \ms{326.96}{0.62} & \ms{69.59}{0.61} & \ms{60.20}{0.30} \\

        \bottomrule
\end{tabular}}
\vspace{-20pt}
\end{table}

The results demonstrate that both components are indispensable. Removing SSA causes clear performance degradation, confirming that the temporal-logical facts it synthesizes provide crucial high-level abstractions for effective reasoning. The w/o CAF variant shows an even greater drop, indicating that without its targeted critique, the LLM-driven evolution lacks guidance. Together, SSA and CAF are essential for enabling TPET to discover specialized, high-performance policies.
\vspace{-5pt}
\subsection{Case Study}
\label{sec:case_study}

\begin{figure}[htb]
        \vspace{-15pt}
	\centering
	\includegraphics[width=\textwidth]{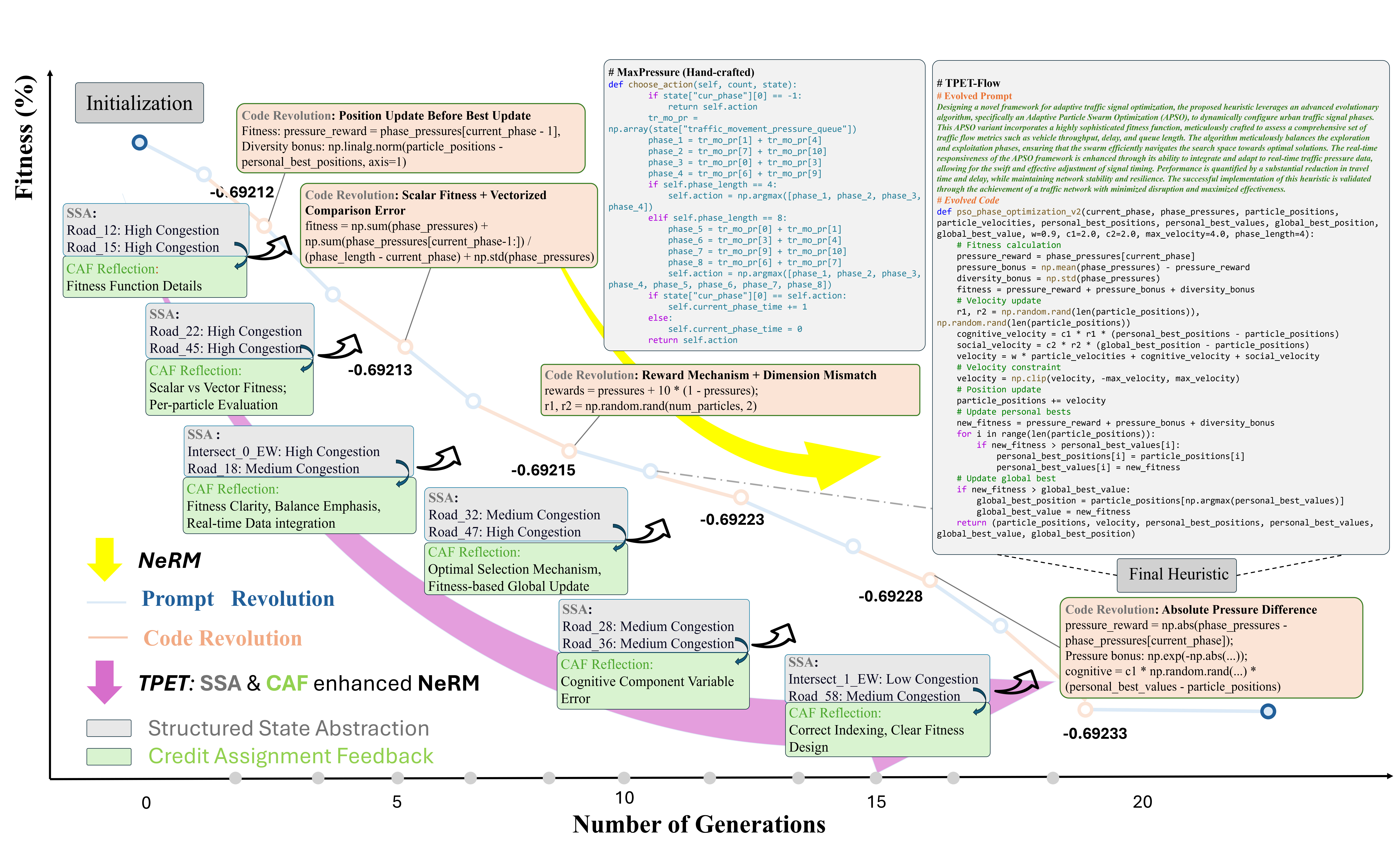}
	\caption{Evolution of \method{} for TSC. We outline the key outputs of the SSA and CAF modules. Moreover,
we present the best algorithm in the final iteration and compare it with Maxpressure. } \label{fig::casestudy}
\vspace{-20pt}
\end{figure}
\vspace{-2pt}

To provide a qualitative understanding of our framework, we visualize the evolutionary discovery path of TPET against the NeRM baseline in Figure~\ref{fig::casestudy}. This figure illustrates the progression from a simple initial policy to a specialized final heuristic, plotting the fitness improvement over generations.

During the evaluation of a policy, the SSA module monitors the runtime statistics. Its primary role is to translate high-dimensional numerical metrics, which are incomprehensible to the LLM, into a set of interpretable, structured facts, such as the high congestion levels on specific roads shown in the figure. This structured information is then passed to the CAF module. CAF aggregates these structured facts from the simulation logs to generate a high-level reflection. This reflection summarizes the policy's defects and guides the main evolutionary direction, for instance, by shifting focus from a simple scalar fitness to a more balanced, multi-faceted one. This structured critique then informs the next code revolution, allowing the LLM to make targeted, intelligent modifications based on concrete, interpretable feedback. This iterative loop of SSA translation and CAF-guided reflection enables TPET to efficiently navigate the search space and discover a robust, multi-layered final heuristic that significantly outperforms the standard MaxPressure baseline.

\section{Conclusion}
In this paper, we presented \method{} framework, an approach utilizes LLM as an evolution engine. Our core contribution lies in architecting a structured feedback loop: the SSA module provides interpretable real-time traffic states, which are then analyzed by the CAF module to generate precise, actionable critiques detailing policy defects. This structured feedback loop empowers the LLM to iteratively debug and refine heuristic policy code, moving beyond opaque optimization towards transparent, explainable, and continually improving control strategies. \method{} demonstrates significant advancements in policy interpretability, development efficiency, and performance over traditional methods. We believe this framework paves a robust path for future research into adaptive, LLM-guided intelligent transportation systems that prioritize both effectiveness and transparency.

%
%
%
\bibliographystyle{splncs04}
\bibliography{mybibliography}

\end{document}